\documentclass[letterpaper, 10 pt, conference]{ieeeconf}  

\IEEEoverridecommandlockouts                              
\makeatletter
\let\NAT@parse\undefined
\makeatother
\usepackage[dvipsnames]{xcolor}

\usepackage{times}
\usepackage{graphicx}
\usepackage{amssymb}
\usepackage{gensymb}
\usepackage{amsmath}
\usepackage{breakurl}

\usepackage{algorithm}
\usepackage{algorithmic}

\usepackage[labelfont={bf},font=small]{caption}
\usepackage[none]{hyphenat}

\usepackage{mathtools, cuted}

\usepackage[noadjust, nobreak]{cite}

\usepackage{tabularx}
\usepackage{amsmath}

\usepackage{float}

\usepackage{pifont}

\newcolumntype{Y}{>{\centering\arraybackslash}X}

\usepackage[]{placeins}

\usepackage{placeins}

\usepackage{tikz}

\usepackage[framemethod=tikz]{mdframed}
\usepackage{diagbox}
\usepackage{afterpage}

\usepackage{stfloats}

\usepackage{atbegshi}
\newcommand{\handlethispage}{}
\newcommand{\discardpagesfromhere}{\let\handlethispage\AtBeginShipoutDiscard}
\newcommand{\keeppagesfromhere}{\let\handlethispage\relax}
\AtBeginShipout{\handlethispage}

\usepackage{comment}
\usepackage{mathtools}

\title{\LARGE \bf
Evolving Order and Chaos: Comparing Particle Swarm Optimization and Genetic Algorithms for Global Coordination of Cellular Automata 
}

\author{Anthony D. Rhodes\\
Portland State University} 

\begin{document}

\maketitle
\thispagestyle{empty}
\pagestyle{empty}

\begin{abstract}
We apply two evolutionary search algorithms: Particle Swarm Optimization (PSO) and Genetic Algorithms (GAs) to the design of Cellular Automata (CA) that can perform computational tasks requiring global coordination. In particular, we compare search efficiency for PSO and GAs applied to both the density classification problem and to the novel generation of "chaotic'' CA. Our work furthermore introduces a new variant of PSO, the Binary Global-Local PSO (BGL-PSO). 
\end{abstract}

\section{INTRODUCTION: CELLULAR AUTOMATA}
Cellular Automata (CA) are discrete, spatially-extended dynamical systems consisting of cells, each of which contains a finite state machine. Given an initial configuration of cells, CA evolve over time by performing computations according to local rules. The input for each local rule is a "neighborhood" of a given cell, and all cells typically use the same local rules -- in this case we say the CA is          \textit{homogeneous}. The space in which the computations of a CA are realized is called the "cellular space" of the CA. The cellular space is typically divided into a 1-d or 2-d lattice structure, although there exist analogous, higher dimensional extensions of CA [1].
\indent CA have been studied extensively as mathematical objects, as models of natural systems, and as architectures for fast parallel computation [2]. They have additionally been applied to a wide range of applications, ranging from particle simulation [3], image processing [4], pattern classification [5], traffic modeling [6], computational biology [7], disease spread [8], crystallization [9], and art [10], among others. 1-d CA in particular represent one of the simplest examples of decentralized systems in which emergent computation can be studied. [11]

\begin{figure}[h]
\begin{tabular}{ll}
\centering
\includegraphics[scale=0.73]{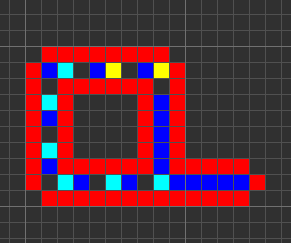}
&
\includegraphics[scale=0.175]{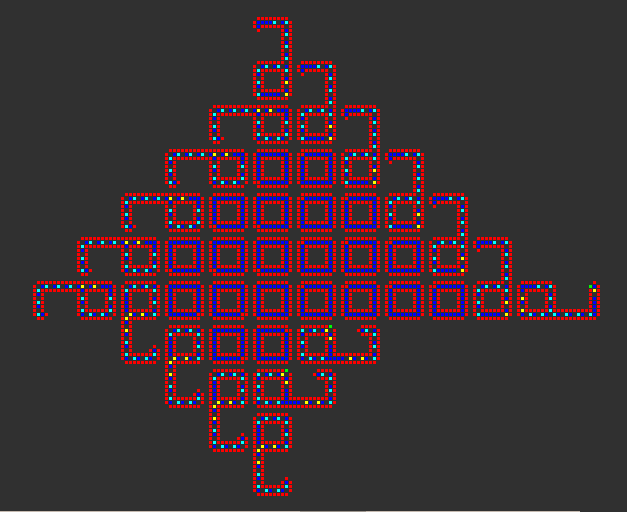}
\end{tabular}

\caption{Left: Langton's Loop, an 86-cell, self-replicating CA.  Right: Colony of Langton's Loops. All figures best viewed in color. }

\label{Fig:Race}
\end{figure}
\subsection{A Brief History Of Cellular Automata}
\indent CA were invented in the 1950s by John von Neumann in an attempt to construct and analyze self-replicating universal computers. von Neumann established universality by proving that an automaton consisting of cells with four orthogonal neighbors and 29 possible states can simulate a Turing machine and self-replicate for a configuration of approximately 100,000 cells [12]. Following von Neumann's work, in 1967 Konrad Zuse conceived of the whole of space as an evolving cellular space, "Rechnender Raum" ("Computing Space") [13]. In 1968 Edgar Codd produced a universal construction with self-replication capabilities requiring only 8 possible states for a configuration of approximately $2 \cdot 10^8$ cells [14]. Abandoning the universality constraint, in 1984 Christopher Langton discovered an 8-cell CA that is capable of self-replication on a configuration of only 86 cells (see "Langton's Loop", Figure 1). Following Codd's results, John Conway [15] developed a breakthrough 2-d CA with a simple rule set meant to mimic high-level organic properties. Conway's so-called "Game of Life" was popularized in Scientific American and was later shown to exhibit properties of universal computation [16]. \\
\indent In an effort to explicate the origins of complexity \textit{in Natura}, Stephen Wolfram devised a taxonomy of 1-d, 2-state CA. He grouped CA into four general rule categories: \textit{homogeneous}, \textit{periodic}, \textit{complex} and \textit{chaotic} structures. In particular, Wolfram conjectured rule 110 (depicted on the cover of his \textit{A New Kind of Science}) is Turing complete (this was proven by Matthew Cook in 2004 [17]) and that rule 30 (see Figure 2) exhibits chaotic or random behavior (this rule is in fact used as a random number generator in the Wolfram language [18]). Wolfram used these and other examples to suggest that the complexity of the universe belies an underlying simplicity in which a few basic rules give rise to complicated and unpredictable behavior [19]. Drawing from Zuse's notion of a Computing Space (Wolfram analogously defines the term "computational universe"), Wolfram also proposed "harvesting" rules that can solve interesting problems. In this way, AI -- particularly \textit{evolutionary methods} -- and complexity theory provide a framework for a novel, largely empirical approach to solving difficult problems by appealing to a comprehensive search in the "computational universe" of programs. 

\begin{figure}[tbp]
\centering
\includegraphics[width=0.97\columnwidth]{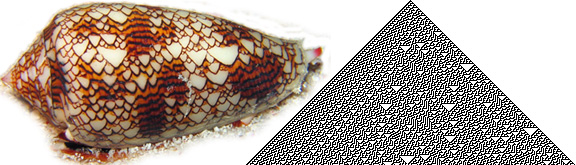}
\caption{Naturally occurring cone snail shell structure (Left) which bears a close resemblance to Wolfram's CA Rule 30 (Right). }
\label{loss_spikes}
\end{figure}
\section{CELLULAR AUTOMATA: FORMAL OVERVIEW}
A CA is a spatial lattice of $N$ cells; a 1-d cellular automaton is a 4-tuple: $A=(k,\Sigma,r,\sigma),$ where $k$ defines the number of discrete states for the CA 
(e.g. $k=2$), $\Sigma$ denotes the alphabet of cell states (e.g. $\Sigma=\{0,1\}$), where  $\Sigma = |k|$; $r$ is the radius of the CA and $\sigma$ represents the CA mapping encoded by the rule set of the CA. As in [19], with the present work we focus on discrete, 1-d, binary CA.

\indent The CA begins with some initial configuration (IC) of cell states. At each time step \textit{t} the CA "evolves" and each cell in the lattice is updated according to the rule set.  Formally, a configuration $s \in \Sigma^{N}$ is a concatenation of cells, namely: $s=s(0)\cdot \cdot \cdot s(N-1)$, where $s(i)$ indicates the state of the ith cell in the configuration. We will denote the state of cell $i$ at time $t$ of the automaton $A$ with initial configuration $s$ by $A_{t,n}(s)$. Cell states are defined in a recursive fashion, so that the state of a particular cell at time $t+1$ depends on the states of its neighbors, as determined by $\sigma:\Sigma^{2r+1}\rightarrow \Sigma$. Hence, $A_{t+1,n}(s)=$ $\sigma(A_{t,n-r}(s)$ ,..., $A_{t,n+r}(s))$.
\indent The (global) configuration of $A(s)$ with IC $s$ at a fixed time $t$ can be thought of as a collection $\{A_{t,i} \}$ over all cells $i \in \{0,...,N-1\}$; we denote this configuration $A_{t}(s)$. The evolution of a particular CA with initial configuration $s$ is therefore written as
follows: $A_{0}(s),A_{1}(s)$,...,$A_{t}(s)$. Oftentimes it is useful to consider a global “snapshot” of the evolution of a particular CA, where the sequence $A_{0}(s)$,$A_{1}(s)$,...,$A_{t}(s)$is realized pictorially, read top-to-bottom, beginning with $A_{0}(s)$, see Figures 3 and 4.\\
\indent We now develop an example of a simple cellular automaton in order to help facilitate the reader's understanding of the previous formalism. Let the cells of our CA be binary ($k=2$, so that: $\Sigma =\{0$ $\equiv$ "white", 1 $\equiv$ $"black"\}$); furthermore, we consider the simple case where $r=1.$ Thus, $\sigma:\Sigma^{3}\rightarrow \Sigma$ and each state transition is determined by a triplet. In following the ordering conventions developed by Wolfram [19], we consider the automaton Rule 250 where the rule set is defined as:
\begin{itemize}
  \item $\sigma(1,1,1)=1$
  \item $\sigma(1,1,0)=1$
  \item $\sigma(1,0,1)=1$
  \item $\sigma(1,0,0)=1$
  \item $\sigma(0,1,1)=1$
  \item $\sigma(0,1,0)=0$
  \item $\sigma(0,0,1)=1$
  \item $\sigma(0,0,0)=0$
  \smash{\raisebox{.5\dimexpr7\baselineskip+0\itemsep+2\parskip}{$\left.\rule{0pt}{.5\dimexpr8\baselineskip+0\itemsep+3\parskip}\right\}\text{$\sigma:\Sigma^{3}\rightarrow \Sigma$}$}} 
  \end{itemize}
\indent As the reader may check, this assignment is realized pictorially by the rule scheme shown in Figure 3. We can see that the sequence of global configurations for this CA is very well-behaved; iterations beginning with the \textit{elementary IC}, $\bar{1}$, are depicted in Figure 4, where: 
    \[ \bar{1}(i) = \begin{cases} 
          1 \text{ if } i = 0 \\
          0 \text{ else} 
       \end{cases}
    \]

\begin{figure}[tbp]
\centering
\includegraphics[width=0.90\columnwidth]{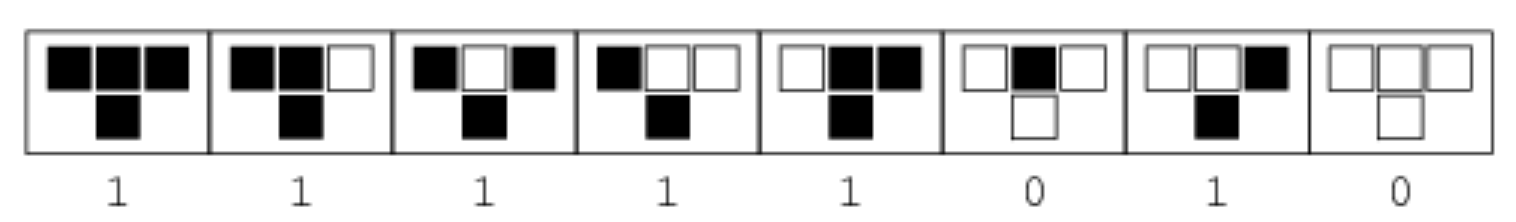}
\caption{Wolfram's Rule 250 rule set. }
\label{loss_spikes}
\end{figure}

\begin{figure}[tbp]
\centering
\includegraphics[width=0.90\columnwidth]{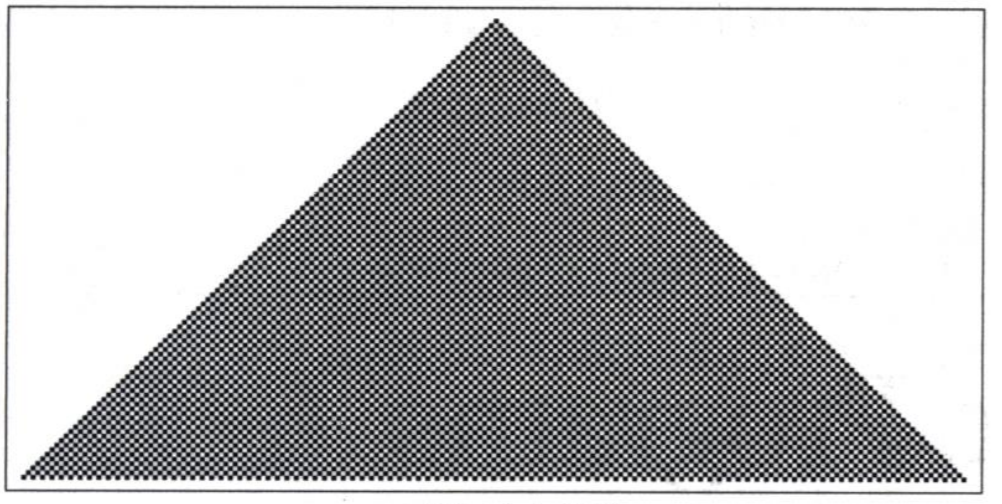}
\caption{CA evolution of Rule 250 for IC $\bar{1}$.}
\label{loss_spikes}
\end{figure}

\subsection{Computational Tasks: Global Coordination of CA}
The current work builds primarily on [20], [21], [22], and [23]. We apply two evolutionary search algorithms, Particle Swarm Optimization and Genetic Algorithms, to two different computational tasks requiring global coordination: (1) the     density classification problem (also called the "majority voting problem") 
for CA and (2) the generation of general "chaotic" (\`a la Wolfram) CA.
\indent For the density classification task, the goal is to find a CA (i.e. to specify a mapping $\sigma$) to determine whether or not an arbitrary IC $s$ has a majority of 1s. In literature this problem is denoted "$\rho_{c}=1/2$", where $c$ represents a \textit{critical} or threshold value for density classification. Concretely, if $\rho_{0}$ is the true proportion of 1s contained in the IC (assume $N$ is odd so that $\rho_{0}$ is unambiguous), and, say, $\rho_{0} > \rho_{c}$, then within a fixed number $M$ timesteps, the CA should generate all 1s, i.e. $A_{t}(s)=1$ $\forall$ $t\geq M.$ 
\indent A solution to the density classification task is trivial for a system with a central controller or global memory. However, designing a solution to the density classification task for a CA or similar system lacking a central controller is non-trivial -- particularly in systems with significant locality constraints (e.g. r $<<$ N). The difficulty of this task is attributable to the fact that the CA must pass information over large distances $(\approx N)$ and additionally process this information across disparate parts of the spatial lattice. Moreover, the density classification tasks requires a system without a central controller to perform computations that are inherently more complex than the computational capacity of each individual cell (or any linear combination of cells) [23]. Density classification is therefore a fitting problem case to test the programmability of "emergent behavior" through the coordination of simple "agents".
\indent In addition to the density classification task, we propose a novel chaotic CA coordination task for which the objective is to take an arbitrary input configuration and generate a maximally "unstructured" output. More specifically, we wish to minimize the compressibility (in an information-theoretic sense) of the output of the CA for any IC. From [24], we augment the \textit{Normalized Compression} (NC) metric so that our objective function is defined as the average of the \textit{piecewise} (i.e. line-by-line output of the CA) NC measure and the \textit{total} (i.e. concatenations of the entire computational history of the CA) NC measure. To this end, we define the \textit{NC piecewise-total} objective function for the chaotic CA coordination task:
\begin{equation}
NC_{p-t}(A(s))=\frac{\sum_{i=0}^{T} K(A_{i}(s))}{T}+K(A(s))
\end{equation}
where A(s) connotes the \textit{total} CA with initial configuration $s$; $A_{i}(s)$ is the ith lattice of the CA output for the IC; $T$ is the final time step (we set $T=150$ for experiments), and K($\cdot$) represents the Kolmogorov complexity of a string, which we approximate with the DEFLATE compression algorithm [25], using a combination of LZSS [26] and Huffman coding [27]. 
\section{Particle Swarm Optimization}
Particle Swarm Optimization is a stochastic, metaheuristic optimization algorithm inspired by the social behavior of animals introduced by Kennedy and Eberhart [20]. The algorithm simulates the motion of "particles" (i.e. organisms) in the search space with the goal of locating a global optimum. Each particle in the "swarm" is influenced by three general factors: the current velocity of the particle, a \textit{social} component, which encodes particle interactions with other particles in the swarm, and a \textit{cognitive} component, which is influenced by the search history of an individual particle. PSO is in general an efficient algorithm that can be applied across large, non-differentiable, continuous or discrete domains [28]. In contrast to GAs (see Section IV), conventional Particle Swarm Optimization algorithms typically avoid the use of genetic operators.  \\
\indent PSO was initially  developed for continuous-valued spaces. In what follows, we first describe the original, continuous PSO algorithm; subsequently,  we adopt a version of PSO for a binary domain appropriate to the previously described CA density classification and chaotic problems.
\subsection{Continuous PSO}
Formally, we wish to optimize a vector-valued function $f:\mathbb{R}^{d}\rightarrow \mathbb{R}$; we call this function $f$ the objective function of our optimization problem. Define the position of the ith particle in the swarm $x_{i}=(x_{i_{1}},...,x_{i_{d}})$, and denote the velocity of the ith particle for the ith particle $v_{i}=(v_{i_{1}},...,v_{i_{d}})$. Furthermore, at each iteration of PSO, we define the \textit{best-visited} state for the ith particle $p_{ibest}=(p_{i_{1}},...,p_{i_{d}})$ and the \textit{global best-visited} position $p_{gbest}=(p_{g_{1}},...,p_{g_{d}})$.The velocity and position update formulas for each particle are given by [22]:
\begin{equation}
\begin{aligned}
&v_i(t+1)= w\cdot v_{i}(t) +  c_{1}\phi_{1}(p_{ibest}-x_{i}(t))+\\ &    \quad \quad  c_{2}\phi_{2}(p_{gbest}-x_{i}(t)) \\
&x_{i}(t+1)=x_{i}(t)+v_{i}(t+1)
\end{aligned}
\end{equation}
where $c_{1}$ and $c_{2}$ are positive constants (we use $c_{1}=c_{2}=2$ [22]); $\phi_{1}$ and $\phi_{2}$ are sampled from $U[0,1]$. In this equation, $w$ is the \textit{inertia weight} which determines the influence of the velocity at the previous time-step. PSO can be sensitive to the choice of inertia weight, as this parameter affects the exploration-exploitation trade-off of the search. Evidence supports the choice of \textit{chaotic inertia weight} [29], defined: $w= (w_{1}-w_{2}) \frac{MaxIterations - t}{MaxIterations}+w_{2} \cdot z$, with $z=4\cdot z_{0}\cdot(1-z_{0}), z_{0} \sim U[0,1],w_{1}=0.4, w_{2}=0.9.$ The basic idea with this particular inertia weight is that $w$ is stochastically increased as a linear sequence beginning at $w_{1}$ and ending at $w_{2}$ over the course of the search; in this manner, this choice of $w$ encourages exploration early in the search.\\

\begin{algorithm}[h]
\caption{Continuous PSO}
\begin{algorithmic}[1]
\FOR{$t$ in \{1,...,$MaxIterations$\}}
    \STATE randomly initialize the position of all the particles $\{x_{i}\}$ in the swarm
    \STATE evaluate the performance of each particle with respect to the objective function: $f(x_{i})$
    \IF{$f(x+{i}) > p_{ibest}$}
     \STATE $p_{ibest} =f(x_{i})$
    \ENDIF
    \IF{$f(x+{i}) > p_{gbest}$}
        \STATE $p_{gbest} =f(x_{i})$
    \ENDIF
    \STATE set $v_i(t+1)= w\cdot v_{i}(t)+c_{1}\phi_{1}(p_{ibest}-x_{i}(t))+  c_{2}\phi_{2}(p_{gbest}-x_{i}(t))$
     \STATE set $x_{i}(t+1)=x_{i}(t)+v_{i}(t+1)$ 
\ENDFOR
\end{algorithmic}
\label{alrc_algorithm}
\end{algorithm}

\subsection{Binary PSO}
Kennedy and Eberhart additionally introduced a binary variation of the PSO alogrithm [20]. The key distcion with binary PSO is that velocities are mapped to probabilities that a particle bit will flip to one. The normalization function they use to compute these probabilities is a standard sigmoid: 

\begin{equation}
v_{ij}'(t) =sig(v_{ij}(t))=\frac{1}{1+e^{v_{ij}(t)}}
\end{equation}
The particle update formula is accordingly defined as follows: 
\begin{equation}
x_{ij}(t+1) \begin{cases} 
      1 \text{ if } r_{ij} < sig(v_{ij}(t))  \\
      0 \text{ otherwise}  \\
       \end{cases}
\end{equation}
where $r_{ij} \sim U[0,1].$ Khanesar \textit{et al}. [22] propose several improvements to the baseline binary PSO algorithm; these changes were further tested and validated through several challenging optimization experiments. Chief among these differences is the interpretation of particle velocity. 
As in the original continuous PSO algorithm, the new binary PSO algorithm maintains velocity as the rate at which the particle changes its bit value (the original Kennedy and Eberhart version of binary PSO velocity in fact affects the exploration-exploitation trade-off in a manner that is antithetical to its intended use, see [22]). In short, Khanesar \textit{et al}. define the (velocity) probability of a change in the jth bit of the ith particle:
\begin{equation}
v_{ij}^{c} \begin{cases} 
      v_{ij}^{1}, \text{ if } x_{ij}=0 \\
      v_{ij}^{0}, \text{ if } x_{ij}=1  \\
       \end{cases}
\end{equation}

These probabilities of a bit flip are defined as functions of the inertia weight, the current velocity and information conveyed through the social and cognitive components of the particle search history, i.e. the global best and personal best states encountered, as described above. The full set of update formulas (see [22] for comprehensive derivation) is given by: 

\begin{equation}
\begin{aligned}
v_{ij}^{1}=wv_{ij}^{1}+d_{ij,1}^{1}+d_{ij,2}^{1} \\
v_{ij}^{0}=wv_{ij}^{0}+d_{ij,1}^{0}+d_{ij,2}^{0}
\end{aligned}
\end{equation}
where: 
\begin{equation}
\begin{aligned}
\text{if } p_{ibest}^{j}=1, \text{ then }d_{ij,1}^{1}=c_{1}r_{1}, d_{ij,1}^{0}=-c_{1}r_{1}\\
\text{if } p_{ibest}^{j}=0, \text{ then }d_{ij,1}^{0}=c_{1}r_{1}, d_{ij,1}^{1}=-c_{1}r_{1}\\
\text{if } p_{gbest}^{j}=1, \text{ then }d_{ij,2}^{1}=c_{2}r_{2}, d_{ij,2}^{0}=-c_{2}r_{2}\\
\text{if } p_{gbest}^{j}=0, \text{ then }d_{ij,2}^{0}=c_{2}r_{2}, d_{ij,2}^{1}=-c_{2}r_{2} 
\end{aligned}
\end{equation}
Here $d_{ij,2}^{0}$ and $d_{ij,2}^{1}$ are temporary values; $c_{1}$ and $c_{2}$ are fixed, as described previously, and $r_{1}, r_{2} \sim U[0,1]$. In summary, formulas (6) and (7) encode stochastic rules which guide the search by encouraging probabilities of a bit flip to cohere with both global and individual search history information. \\
\indent Finally, the next particle state is computed as: 
\begin{equation}
\begin{aligned}
 x_{ij}(t+1)= \begin{cases} 
      x_{ij}(t)+1 \text{ mod 2 if } r_{ij} < v_{ij}'(t) \\
      x_{ij}(t) \text{ if } r_{ij} > v_{ij}'(t) \\
       \end{cases}
\end{aligned}
\end{equation}
where $r_{ij} \sim U[0,1]$. We summarize the Binary PSO algorithm below.

\begin{algorithm}[h]
\caption{Binary PSO}
\begin{algorithmic}[1]
\FOR{$t$ in \{1,...,$MaxIterations$\}}
    \STATE randomly initialize the position of all the particles $\{x_{i}\}$ in the swarm
    \STATE evaluate the performance of each particle with respect to the objective function: $f(x_{i})$
    \IF{$f(x+{i}) > p_{ibest}$}
     \STATE $p_{ibest} =f(x_{i})$
    \ENDIF
    \IF{$f(x+{i}) > p_{gbest}$}
        \STATE $p_{gbest} =f(x_{i})$
    \ENDIF
    \STATE update velocity of particles according to formulas (6) and (7) 
    \STATE calculate the change bits velocity, $v_{ij}^{c}$
    \STATE generate $r_{ij} \sim U[0,1]$; update position of particle using (8)
\ENDFOR
\end{algorithmic}
\label{alrc_algorithm}
\end{algorithm}

\subsection{Binary Global-Local PSO}
Despite the efficiencies introduced in both the original PSO and binary PSO algorithms, these techniques are still nevertheless susceptible to converging to local extrema. This issue is unfortunately compounded in the current problem domain of searching for configurations of CA that solve the density and chaotic problems, as these search spaces are both sparse and highly discontinuous. To remedy some of the deficiencies of PSO and binary PSO in this space, we introduce a novel variant of PSO, the Binary Global-Local PSO.\\
\indent The key changes that we introduce in the BGL-PSO algorithm include the addition of swarm "locality" via neighborhoods. For the duration the BGL-PSO algorithm, we define a topological structure across the swarm. This structure gives rise to "neighborhoods", where each particle is assigned a \textit{fixed} index in $\mathbb{Z}/N=\{0,...,N-1\},$ where $N$ indicates the size of the swarm, $S$. 
The neighborhood of a particle $x_{i}$ is then defined: $Nbh(x_{i})=\{x_{j}\in S \text{ s.t. } |i-j|\leq \Delta_{Nbh} \}$, where $\Delta_{Nbh}$ is a neighborhood size tolerance parameter. \\
\indent In addition to the social and cognitive components present in the original PSO algorithm, we introduce the notion of a topological neighborhood that yields a \textit{local-social} parameter, denoted $p_{ls}$. We leverage this parameter in the course of a search to encourage the swarm to perform several local searches in parallel, whilst still receiving information from a "global" source encoded via $p_{gbest}$. We furthermore include a mutation process in our algorithm to help address the severe discontinuities present in the CA search space problem domain. For each particle update step in our algorithm, we accordingly sample from a \textit{mutation pmf} that determines the number of bits to randomly flip (we use a pmf that is strongly right-tailed so as to maintain a non-aggressive mutation procedure). \\
\indent We use the following update formulas for the BGL-POS algorithm: 

\begin{equation}
\begin{aligned}
v_{ij}^{1}=wv_{ij}^{1}+d_{ij,1}^{1}+(d_{ij,2}^{1}+d_{ij,3}^{1})\div 2 \\
v_{ij}^{0}=wv_{ij}^{0}+d_{ij,1}^{0}+(d_{ij,2}^{0}+d_{ij,3}^{0})\div 2
\end{aligned}
\end{equation}
here $p_{ils}^{j}$ represents the jth bit of $max(f(Nbh(x_{i})))$; if $p_{ils}^{j} =1,$ then $d_{ij,3}^{1}=c_{2}r_{3}, d_{ij,3}^{0}=-c_{2}r_{3}$ and if $p_{ils}^{j} =0,$ then $d_{ij,3}^{0}=c_{2}r_{3}, d_{ij,3}^{1}=-c_{2}r_{3}$, for $r_{3} \sim U[0,1]$.\\
\indent We describe the BGL-POS in full below. 
\begin{algorithm}[h]
\caption{BGL-PSO}
\begin{algorithmic}[1]
\FOR{$t$ in \{1,...,$MaxIterations$\}}
    \STATE randomly initialize the position of all the particles $\{x_{i}\}$ in the swarm
    \STATE evaluate the performance of each particle with respect to the objective function: $f(x_{i})$
    \IF{$f(x+{i}) > p_{ibest}$}
     \STATE $p_{ibest} =f(x_{i})$
    \ENDIF
    \IF{$f(x+{i}) > p_{gbest}$}
        \STATE $p_{gbest} =f(x_{i})$
    \ENDIF
    \IF{$max(f(Nbh(x_{i}))) > p_{ls}$}
        \STATE $p_{ls} =max(f(Nbh(x_{i})))$
    \ENDIF
    \STATE update velocity of particles according to formula (9) 
    \STATE calculate the change bits velocity, $v_{ij}^{c}$
    \STATE generate $r_{ij} \sim U[0,1]$; update position of particle using (8)
    \STATE sample from mutation pmf; flip bits of particle accordingly
\ENDFOR
\end{algorithmic}
\label{alrc_algorithm}
\end{algorithm}

\section{Genetic Algorithms}
For comparison, we likewise apply GAs to the task of evolving CA for density estimation and chaos generation. The purpose of comparing PSO and GAs for these CA-related tasks is not to determine whether one approach is incontrovertibly superior; instead, we wish to highlight their (sometimes nuanced) differences, and in particular to offer a potential improvement to PSO in the challenging domain of large, sparse search spaces with a high degree of discontinuity and non-convexity. Relative to PSO, GAs are a  well-known algorithm archetype; for readers interested in a  comprehensive treatment of GAs, we recommend: [30,31]. We henceforth assume the reader has a basic understanding of the GA paradigm, which is summarized below. \\
\indent Following [21,23] we apply a GA for the $\rho_{c}=1/2$ task with $r=3$ (the CA neighborhood rule set size), which yields a search space of size $2^{128}$, which is prohibitive for exhaustive search. In addition, we let $N=149$ (the width of the CA input), and, as previously mentioned, we set $T=150$ (the maximum number of CA computation steps). We additionally apply these same parameter settings to the chaos generation problem.\\
\indent The GA population is initialized as a set of 100 random binary strings of length 128. Again, following [23], the fitness rule $F_{100}$ is defined as the fraction of 100 randomly chosen ICs from the "flat" uniform distribution $\rho \sim U[0,1]$ (the ICs are randomly sampled anew each generation) which were "classified" correctly. We remind the reader that classification entails the following:  if $\rho_{0}$ is the true proportion of 1s contained in the IC,  and $\rho_{0}> \rho_{c}$,  then within $T$ computation steps, the CA should generate all 1s. The justification for using the "flat" uniform distribution in place of random distribution sampled independently for each bit, is that the latter distribution will be biased, since it is heavily weighted at $\rho = 1/2$; in this way the flat uniform distribution is a more robust representation of the variation inherent to the density classification problem. After evaluating each CA in the initial population using $F_{100}$, we retain (without modification) the top $20\%$ "elite" group in the population; we create the remainder of the new offspring using single-point crossover using samples (with replacement) of chromosomes pooled uniformly from this elite group. These offspring are each mutated at exactly two randomly chosen positions. This process was repeated for 200 epochs. 
\begin{algorithm}[h]
\caption{GA}
\begin{algorithmic}[1]
\FOR{$t$ in \{1,...,$MaxIterations$\}}
    \STATE randomly initialize the initial population of binary strings (initial chromosomes)
    \STATE sample 100 ICs from the "flat" uniform distribution: $\rho \sim U[0,1]$
    \STATE evaluate the performance of each chromosome using $F_{100}$
    \STATE retain top $20\%$ of the elites from population; apply single-point crossover from elite group using sampling with replacement
    \STATE apply mutation to each offspring at exactly two randomly chosen positions 
\ENDFOR
\end{algorithmic}
\label{alrc_algorithm}
\end{algorithm}

\section{EXPERIMENTAL RESULTS}
We apply binary PSO, BGL-PSO and GAs to the CA density estimation and chaos generation tasks with common parameter settings $r=3$, $N=149$, $T=150$, over $200$ epochs, across $10$ independent trials in each case. Both PSO algorithms utilize chaotic inertia weight schemes; we set $c_{1}=c_{2}=2$ as is conventional for PSO; we furthermore set $\Delta_{Nbh}=5$ for BGL-PSO. The mean of the best fitness values for each method ($F_{100}$ and $NC_{p-t}$, respectively) at the conclusion of the search following 200 epochs is shown in Table 1. \\
\indent Given the difficulty of the search problem, each algorithm performed well in general. However, in each case the GA-based search outperformed both the binary PSO and BGL-PSO algorithms. Figures 6 and 7 demonstrate results of a typical search for each task for the GA algorithm. Figures 7 and 8 show CA output on the best-performing chromosome on each task generated by GA search. We conjecture that GA search dominated PSO-based methods for the given tasks due in part to the non-contiguous "jumps" engendered by the genetic crossover operation, and that this non-contiguous behavior is beneficial current large, non-convex search space. While PSO was -- absent aggressive non-local jumping -- more prone to be caught in local extrema, many times these local extrema were nonetheless found relatively early in the search with PSO; moreover, although the GA was typically more effective at finding a single best candidate solution, the PSO algorithm by contrast commonly found more instances of \textit{different} quality solutions. Our BGL-PSO algorithm also consistently outperformed the binary PSO algorithm for the given CA-related tasks. 

\begin{table}
\centering
\begin{tabular}{|l|c|c|}\hline

   & Density ($F_{100}$)& Chaos ($NC_{p-t}$)   \\ \hline
Binary PSO & 0.51 & 0.773410  \\ \hline
BGL-PSO & 0.62 & 0.777963   \\ \hline
GA & 0.90 & 0.779528 \\ \hline
\end{tabular}
\caption{Summary of mean fitness over 10 trials at 200 epochs each of best candidate for Binary PSO, BGL-PSO and GA search algorithms applied to density classification and chaos generation CA-based tasks. } 
\end{table}
\begin{figure}[tbp]
\centering
\includegraphics[width=0.95\columnwidth]{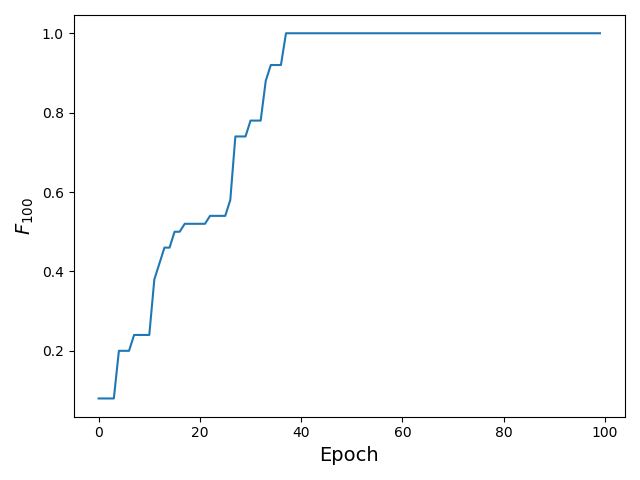}
\caption{Representative GA search result plot for density classification task showing best fitness score ($F_{100}$) per epoch.}
\label{loss_spikes}
\end{figure}

\begin{figure}[tbp]
\centering
\includegraphics[width=0.95\columnwidth]{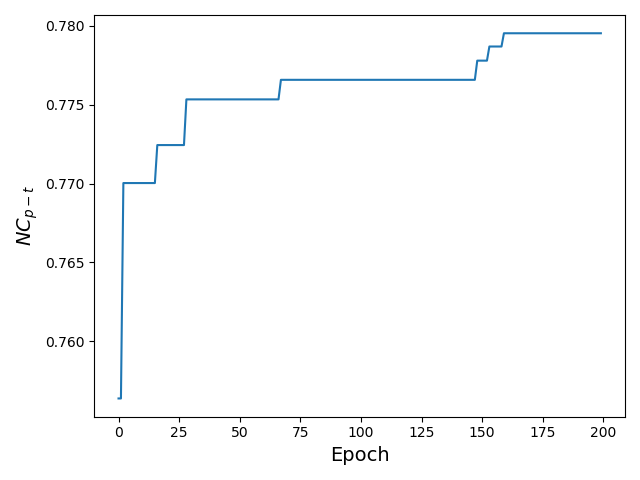}
\caption{Representative GA search result plot for chaos generation showing best fitness score ($NC_{p-t}$) per epoch.}
\label{loss_spikes}
\end{figure}

\begin{figure}[tbp]
\centering
\includegraphics[width=1.1\columnwidth]{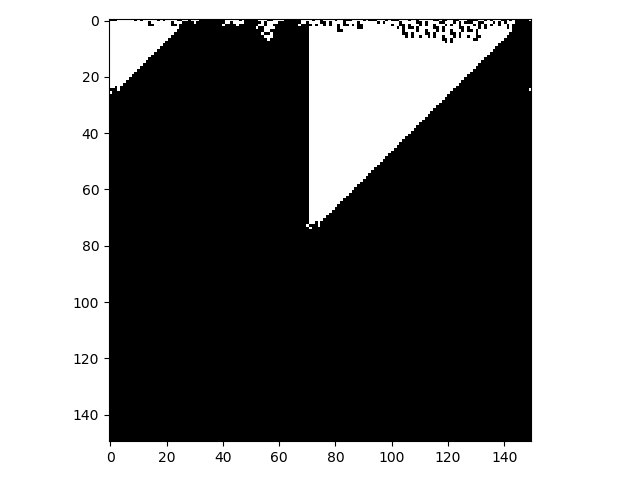}
\caption{Example CA computation for GA-based candidate on density classification task ($F_{100}=1.00$), given random IC with $\rho_{0}<0.5$.}
\label{loss_spikes}
\end{figure}
\begin{figure}[tbp]
\centering
\includegraphics[width=1.1\columnwidth]{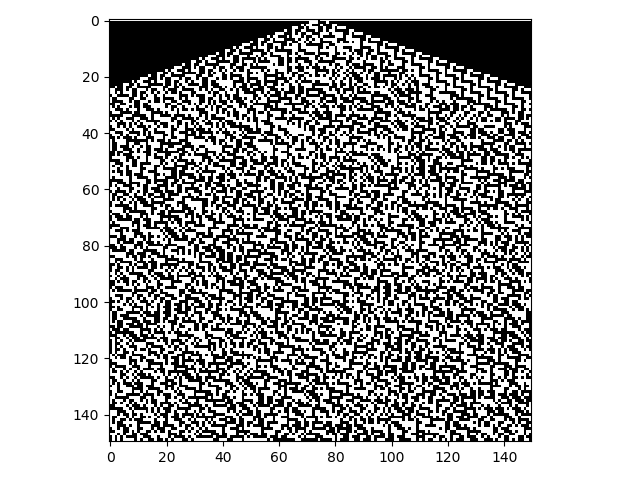}
\caption{Example CA computation for GA-based candidate on chaos generation task ($NC_{p-t}=0.781003$) given IC $\bar{0}$.}
\label{loss_spikes}
\end{figure}
\section{References} 
[1] Druon, S. et al. Efficient cellular automata for 2D / 3D free-form modeling. Journal of WSCG. 2003, vol. 11, no. 1-3.

[2] Bandini, S. et al. Cellular Automata: From a Theoretical Parallel Computational Model to its Applications to Complex Systems. Parallel Computing 27 (2001) 539-553.

[3] Burstedde, C. et al. Simulation of Pedestrian Dynamics Using a 2-D Cellular Automaton. Physica A: Statistical Mechanics and its Applications
Volume 295, Issues 3–4, 15 June 2001, Pages 507-525.

[4] Adriana Popovici and Dan Popovici. Cellular Automata in Image Processing. AMS 2000 MSC: 68Q80, 68U10.

[5] Pradipta Maji, Chandrama Shaw, Niloy Ganguly, Biplab K. Sikdar, and P. Pal Chaudhuri. 2003. Theory and Application of Cellular Automata For Pattern Classification. Fundam. Inf. 58, 3-4 (August 2003), 321-354.

[6] Rickert, M. et al. Two lane traffic simulations using cellular automata. Physica A: Statistical Mechanics and its Applications
Volume 231, Issue 4, 1 October 1996, Pages 534-550.

[7] Hwang, Minki et al. Rule-Based Simulation of Multi-Cellular Biological Systems—A Review of Modeling Techniques. Cell Mol Bioeng. 2009 Sep; 2(3): 285–294.

[8] Sirakoulis, G. Ch. et al. A cellular automaton model for the effects of population movement and vaccination on epidemic propagation. Ecological Modelling 133 (0000) 209–223.

[9] Kazuhiro Hikami1, Rei Inoue1, and Yasushi Komori. Crystallization of the Bogoyavlensky Lattice. J. Phys. Soc. Jpn. 68, pp. 2234-2240 (1999). 

[10] Hoffman, Peter. Towards an "automated art": Algorithmic processes in Xenakis' compositions. Contemporary Music Review, 21:2-3, 121-131. 

[11] Mitchell, Melanie, Peter T. Hraber, and James P. Crutchfield. "Revisiting the Edge of Chaos: Evolving Cellular Automata to Perform Computations." SFI Working Paper 1993-03-014 (1993).

[12] Von Neumann, J. and A. W. Burks (1966). Theory of self-reproducing automata. Urbana, University of Illinois Press.

[13] Konrad Zuse, 1969. Rechnender Raum. Braunschweig: Friedrich Vieweg and Sohn. 

[14] Codd, Edgar F. (1968). Cellular Automata. Academic Press, New York.

[15] Games, M. (1970). The fantastic combinations of John Conway’s new solitaire game “life” by Martin Gardner. Scientific American, 223, 120–123.

[16] Rendell, Paul. A Universal Turing Machine in Conway's Game of Life. International Conference on High Performance Computing and Simulation (2011), pp 764-772. 

[17] Cook, Matthew (2004). "Universality in Elementary Cellular Automata." Complex Systems. 15: 1–40.

[18] Sipper, Moshe; Tomassini, Marco (1996). "Generating parallel random number generators by cellular programming". International Journal of Modern Physics C. 7 (2): 181–190. 

[19] 2002. A New Kind of Science. Wolfram Media Inc., Champaign, Ilinois, US, United States.

[20] James Kennedy and Russell Eberhart. Particle Swarm Optimization. Neural Networks, 1995. Proceedings., IEEE International Conference. 

[21] Das, Rajarshi, P. James, Melanie Mitchell, and James E. Hanson. "Evolving Globally Synchronized Cellular Automata." SFI Working
Paper 1995-01-005 (1995).

[22] Khanesar, Mojtaba et al. A novel binary particle swarm optimization.  Mediterranean Conference on Control and Automation. (2007). 

[23] Mitchell, Melanie et al. Evolving Cellular Automata with Genetic Algorithms:
A Review of Recent Work. Proceedings of the First International Conference on Evolutionary Computation and Its Applications (EVCA). (1996). 

[24] Zenil, Hector. Asymptotic Behavior and Ratios of Complexity in Cellular Automata. International Journal of Bifurcation and Chaos. (2013).

[25] L. Peter Deutsch (May 1996). DEFLATE Compressed Data Format Specification version 1.3. IETF. p. 1. sec. Abstract. 

[26] Storer, James A.; Szymanski, Thomas G. (October 1982). "Data Compression via Textual Substitution". Journal of the ACM. 29 (4): 928–951.

[27] Huffman, D. (1952). "A Method for the Construction of Minimum-Redundancy Codes" (PDF). Proceedings of the IRE. 40 (9): 1098–1101. \\

[28] James Kennedy and Russell Eberhart. A Discrete Binary Version Of The Particle Swarm Algorithm - Systems, Man, and Cybernetics, 1997. 'Computational Cybernetics and Simulation'. (1997).

[29] Feng, Yong et al. Chaotic Inertia Weight in Particle Swarm Optimization. Second International Conference on Innovative Computing, Informatio and Control (ICICIC 2007).

[30] Holland, John (1992). Adaptation in Natural and Artificial Systems. Cambridge, MA: MIT Press.

[31] Mitchell, Melanie (1996). An Introduction to Genetic Algorithms. Cambridge, MA: MIT Press.

\clearpage

\end{document}